\documentclass[letterpaper, 10 pt, conference, twocolumn]{ieeeconf}

\IEEEoverridecommandlockouts
\usepackage{cite}
\usepackage{algorithm}
\usepackage{algorithmic}
\usepackage{ctable}
\usepackage{graphicx}
\usepackage{multicol}
\usepackage{multirow}
\usepackage{textcomp}

\graphicspath{{figures/}}

\usepackage{soul}
\let\labelindent\relax
\usepackage{enumitem}

\def\BibTeX{{\rm B\kern-.05em{\sc i\kern-.025em b}\kern-.08em T\kern-.1667em\lower.7ex\hbox{E}\kern-.125emX}} 

\definecolor{deep-red}{RGB}{192, 0, 0}
\definecolor{deep-purple}{RGB}{120, 0, 170}
\definecolor{good-green}{RGB}{0,175,0} 
\definecolor{purple}{RGB}{210, 0, 210}

\makeatletter
\let\NAT@parse\undefined
\makeatother
\usepackage[colorlinks]{hyperref}
\hypersetup{
    colorlinks=true,
    linkcolor=magenta,
    citecolor=blue,
    filecolor=magenta,      
    urlcolor=cyan,
    pdfpagemode=FullScreen,
}

\begin{document}

\title{A Digital Twin for Telesurgery under Intermittent Communication}
\author{%
     Junxiang Wang$^{*2}$, Juan Antonio Barragan$^{*1}$, Hisashi Ishida$^{*1}$, \\ Jingkai Guo$^1$, Yu-Chun Ku$^1$, Peter Kazanzides$^1$%
     \thanks{$^1$Department of Computer Science, Johns Hopkins University, Baltimore, MD, 21218, USA.
     Email: {\tt \{jbarrag3, hishida3, jguo90, yku4, pkaz\}@jhu.edu}}
     \thanks{$^2$Robotics Institute, Carnegie Mellon University, Pittsburgh, PA, 15213, USA.
     Email: {\tt junxiang@cmu.edu}}
     \thanks{$^*$Equal contributions}
}

\maketitle

\begin{abstract}
Telesurgery is an effective way to deliver service from expert surgeons to areas without immediate access to specialized resources. However, many of these areas, such as rural districts or battlefields, might be subject to different problems in communication, especially latency and intermittent periods of communication outage. This challenge motivates the use of a digital twin for the surgical system, where a simulation would mirror the robot hardware and surgical environment in the real world. The surgeon would then be able to interact with the digital twin during communication outage, followed by a recovery strategy on the real robot upon reestablishing communication. This paper builds the digital twin for the da Vinci surgical robot, with a buffering and replay strategy that reduces the mean task completion time by 23\% when compared to the baseline, for a peg transfer task subject to intermittent communication outage. The relevant code can be found here: \url{https://github.com/LCSR-CIIS/dvrk_digital_twin_teleoperation}.
\end{abstract}

\section{Introduction}
\label{sec:intro}
Telesurgery, the process of conducting surgeries remotely with the assistance of robotic systems, has been gaining increasing attention. Performing surgical procedures in this remote fashion would especially benefit less privileged areas without convenient access to specialized surgical resources \cite{shahzad2019telesurgery,Heemeyer2025}. Telesurgery addresses the issue of the disproportionately low number of expert surgeons compared to the overall population size in low- to medium-income countries \cite{holmer2015global, mehta2022embracing}. It can also reduce the cost of travel for both patients and medical professionals \cite{mohan2021telesurgery, singh2022telesurgery}, as well as facilitating the training of prospective and less experienced surgeons \cite{hung2018telementoring}.

The da Vinci Surgical System (Intuitive Surgical, Sunnyvale, USA) has achieved widespread clinical adoption,
with over 8,600 systems across the world, over 14 million procedures performed since inception, and over 76 thousand surgeons trained on the system (through October 2023) \cite{intuitive2023esg}.
Although the da Vinci does not currently support remote telesurgery, it presents a canonical model for telesurgery in which a surgeon sits at a console, away from the sterile field, and teleoperates robotic instruments inside the patient.

As early as 2009, Garcia \emph{et al.} \cite{garcia2009trauma} proposed Trauma Pod, a solution for both teleoperated surgeries and autonomous supply maintenance for military applications, where the teleoperation component is handled with a modified da Vinci system, allowing surgeons to be distanced from the battlefield while performing time-critical procedures on the wounded. As part of the NASA Extreme Environment Mission Operation (NEEMO) project, another surgical robotics system, Raven, was involved in a pursuit to deliver telesurgery to the bottom of the ocean \cite{takacs2016origins, haidegger2011surgery}.
These examples demonstrate the wide range of applications that would benefit from telesurgery, where medical expertise can be delivered to anywhere on Earth, all with the surgeon located at their usual working area, but teleoperating a robot across a long distance.

When the first telesurgery was conducted in 2001 \cite{marescaux2002transcontinental}, crossing the borders of continents to operate between France and the USA, a special terrestrial network based on asynchronous transfer mode technology was required to maintain a low time lag for telecommunication. With more recent developments in 5G networks, fast communication between any two destinations on Earth has never been as readily available as it is now \cite{dohler2021internet}. The ease of telecommunication has driven increasing attempts at carrying out telesurgeries across distance \cite{rocco2024insights}, which raises important questions regarding the stability of communication during the operation. As mentioned before, a major goal for telesurgery is to allow less privileged areas to receive specialized operations from elite surgeons. It would only be logical to assume that the quality of communication throughout the surgery would not remain constant, considering that these areas are likely rural, or in the case of battlefields, may be subject to constant disturbances. As a result, there might be disruptions in communication in the form of increased delay, or periods of communication failure. In this paper, we focus on the latter issue, where during the surgery, there are short periods of communication outage between the operator and the teleoperated device.

While some existing solutions to the communication loss problem include having multiple surgeon consoles to ensure that at least one console would maintain communication \cite{rocco2024insights}, approaches like this would interrupt the surgeon's experience as they switch to a different console to continue the operation. Multiple connections might also increase the communication burden from the remote side, hence leading to more communication degradations.

\begin{figure*}[ht]
    \vspace{3mm}
    \centering
    \includegraphics[width=0.96\linewidth]{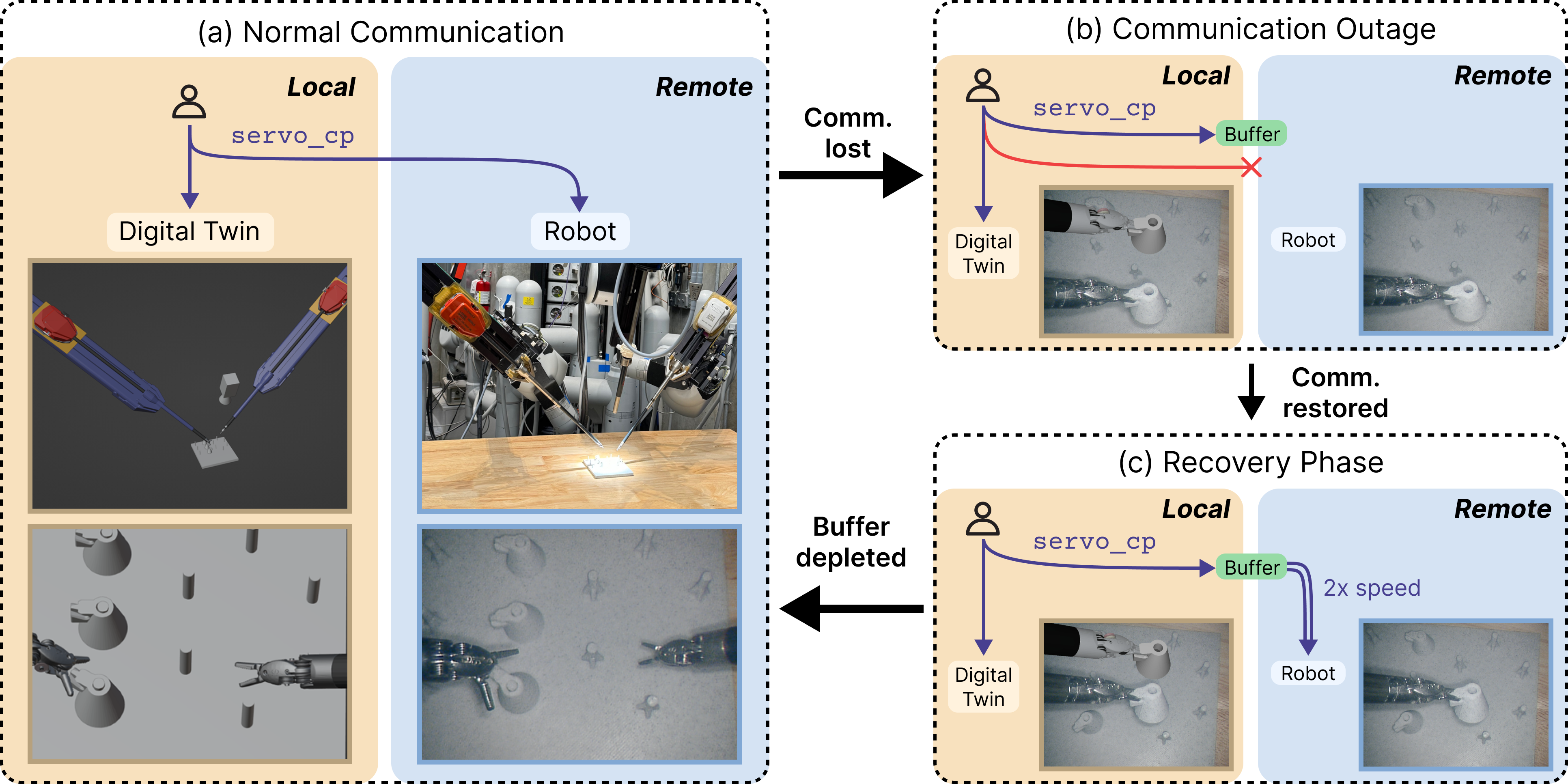}
    \caption{Our proposed digital twin framework for the da Vinci Research Kit to mitigate task disruptions due to communication losses in telesurgery applications. (a) During normal communication, the user sends control commands to both the real and digital twin robots. (b) During a communication outage, the user continues the task using the digital twin robot to minimize task disruption. (c) When communication is restored, the remote robot catches up with the commands of the digital twin.}
    \label{fig:main_figure}
\end{figure*}

In contrast, our approach strives to build a smooth experience for the operator, where we create a digital twin of the surgical environment in simulation, and when a communication outage occurs, the user would instead command the virtual instrument. Our approach draws inspiration from the field of model-mediated teleoperation \cite{Mitra2008}, in which the user interacts with a virtual simulation environment that acts as a medium between the teleoperator and the remote robot. The instrument from our digital twin would be displayed as an augmented reality (AR) overlay on top of the real endoscopic image, which would be frozen during communication outage. The surgeon's input trajectory to the virtual instrument would be saved and subsequently replayed at a faster speed onto the real remote robot when communication is restored (see Fig.~\ref{fig:main_figure}). Our approach builds on a previous work \cite{ishida2023semi}, where preliminary methods of recovering from communication loss were tested in a purely virtual simulation. The main contributions of this work are as follows:
\begin{itemize}[leftmargin=*]
    \item Construction of a digital twin of the da Vinci system, which is registered to the real system with sufficient accuracy for AR overlay.
    \item A method to allow the surgeon to operate on the digital twin during communication outage, with changes replayed on the remote system when communication is restored.
    \item A user study to evaluate the performance of the proposed method with respect to the baseline case where teleoperation is disabled during communication outage.
\end{itemize}
The proposed replay method is a first step toward the application of digital twin technology to compensate for communication challenges during remote teleoperation, and we hope that our system will enable other researchers to test more advanced recovery strategies. In particular, our system enables easy switching of the input device, so one may use a small tabletop manipulator to control the surgical instruments instead of the traditional surgeon console. Additionally, the system also allows customization of the pattern of communication issues. As a result, we encourage collaborative effort from the community to explore different input devices, different communication imperfections including time delays, as well as different control strategies to overcome poor communication channels, so that telesurgery can see broader and safer applications around the world.

\section{Related works}
\label{sec:related}

Related work consists primarily of augmented reality (AR) overlays, especially when the AR overlay is derived from a digital twin of the surgical scene, as well as approaches such as model-mediated teleoperation that are often applied in scenarios with large communication delays.

Some of the largest surgical application areas of AR are orthopedic surgeries and maxillofacial surgeries \cite{lex2023clinical} \cite{barcali2022augmented}. For instance, Dennler \emph{et al.} \cite{dennler2021augmented} conducted a study where during an orthopedic surgery, the patient's computed tomography (CT) scans were displayed through an AR headset, and the surgeon could reposition the CT scan overlay to increase situational awareness.
As another example, Liu \emph{et al.} \cite{liu2021augmented} considers orbital floor reconstruction surgery, where a preoperative CT model of the patient's orbital anatomy is shown as an AR overlay, aligned with the patient's body during the operation. In these surgeries, as well as brain and spinal cord surgeries, AR overlays primarily bring in the benefit of fewer attention shifts, no interruptions due to loss of line of sight, as well as less exposure to radiation \cite{avrumova2023augmented}. A more in-depth summary of the use of AR in surgical robotics applications can be found in \cite{qian2019review}.

While these works present important applications of AR in surgeries, digital twins, like the one presented in this paper, extend far beyond the ability of basic overlays. By mirroring the configuration of the real world in a virtual environment, digital twins can utilize privileged information in simulation to compensate for imperfections in the real setup, which can be particularly crucial in a surgical environment. The imperfection we focus on in our work is intermittent communication outages in telesurgery.


Multiple studies have reported the presence of time delays in long-distance teleoperation for surgical robotics \cite{Arata2006remote, Lum2009Teleoperation}, which is closely related to the communication outage problem under consideration. Lum \emph{et al.} demonstrated that even a latency of 0.25 to 0.5 seconds can double the task completion time in teleoperated surgery \cite{Lum2009Teleoperation}. Teleoperation subject to varying amounts of latency presents challenges analogous to those encountered in communication outages, and many methods have been developed for solving latency issues\cite{orosco2021compensatory, UDDIN201682}, \cite{UDDIN201682}, \cite{richter2019augmented}. This is especially common in bilateral teleoperation, which includes haptic feedback to the operator, because latency tends to destabilize bilateral control systems.

The work by Richter \emph{et al.} \cite{richter2019augmented} bears the most resemblance to ours, since they use a virtual environment to assist in overcoming communication latency when teleoperating a da Vinci system. The authors developed a system where a velocity-based kinematic extrapolation is used to display the position of the instrument a short time period into the future as a virtual overlay, with positioning accuracy mediated by an Extended Kalman Filter. Our work draws key distinctions from \cite{richter2019augmented} as: (1) we focus on communication outage, rather than communication latency, and (2) we do not use a geometry-based visualization, but instead build a digital twin based on a physics simulation, which would allow interactions with other objects in the environment to be computed from dynamics, even during periods of communication outage.

The most relevant approaches that can also apply to communication outage, are those that rely on a model of the remote environment, such as model-mediated telemanipulation \cite{Mitra2008} where the operator directly interacts with a model that is updated by feedback from the remote system. Similarly, in teleprogramming \cite{Funda1992} and tele-sensor-programming \cite{Hirzinger1993}, the operator interacts with a simulated remote environment and teleprograms the remote robot through a sequence of elementary motion commands.
Our approach differs in that the operator interacts with the remote environment during normal communication, and the system automatically switches to the digital twin (simulation model) during communication outage.

\section{Methodology}
\label{sec:method}

Our proposed framework involves three main components: the remote robot, the local user, and digital twin system of the robot and the environment. The digital twin's goal is to enable the user to perform their task with minimal disruptions when faced with communication outages. The efficacy of the overall framework significantly depends on carefully calibrating the digital twin so that it closely tracks the remote robot and environment during periods of normal communication and behaves realistically during periods of communication outage. Once communication is restored, a recovery phase is necessary to reconcile the remote robot's state with the digital twin. In this paper, we have opted for a simple replay strategy in which the robot catches up with the digital twin. However, more advanced recovery strategies can be easily implemented with the proposed framework. Fig.~\ref{fig:main_figure} shows an overview of our system in a full cycle of normal communication, communication outage, recovery phase, and finally back to normal communication. 

The remaining sections in the methodology are divided as follows. The hardware and software components used to build our framework are explained in Sec. \ref{sec:hardware}. The construction and registration of the digital twin are explained in greater detail in Sec.~\ref{sec:dt}. The control logic of our system is explained in Sec.~\ref{sec:control}. Lastly, software implementation details of the digital twin are explained in Sec.~\ref{sec:ar}.

\subsection{Hardware and software components} \label{sec:hardware}
Our system is built upon the da Vinci Classic, powered by open-source controller hardware and software developed in the da Vinci Research Kit (dVRK) \cite{kazanzides_open-source_2014}. The setup consists of two Master Tool Manipulators (MTMs) and two Patient Side Manipulators (PSMs) with large needle driver instruments mounted on them. The former serves as the teleoperation device for our framework and the latter as the remote teleoperated robot. Lastly, the dVRK has an endoscopic camera mounted on an Endoscopic Camera Manipulator (ECM) to stream images from the remote side back to the user. As a dynamic simulation environment for our digital twin, we utilize the Asynchronous Multi-Body Framework (AMBF) \cite{MunawarAMBF}  due to its tight integration with the dVRK system.


\subsection{Digital twin models and calibrations}
\label{sec:dt}

\begin{figure}[!b]
    \centering
    \includegraphics[width=0.7\linewidth]{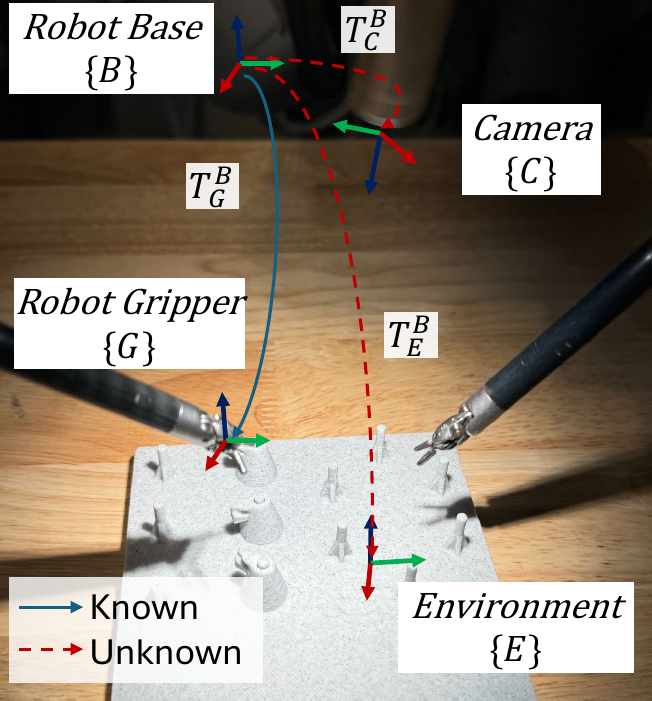}
    \caption{Transformations between different components of the digital twin system, separated into ones that are known and unknown (requiring calibrations).}
    \label{fig:transform}
\end{figure}

To create a highly accurate and interactive digital twin framework, it is necessary to carefully model the remote robot, cameras, and the environment, as well as the spatial relationship between these components (see Fig. \ref{fig:transform}). An accurate model of the PSM and the large needle driver instrument was obtained from the simulation environment in \cite{munawarOpen2022}.

To create a realistic virtual camera, intrinsic and extrinsic parameters from the endoscopic camera of the dVRK system were estimated. Intrinsic camera parameters, such as field of view (FoV), focus, and distortion parameters, were calculated using the Robot Operating System (ROS) image pipeline calibration tools. The extrinsic camera parameters in our framework refer to the rigid transformation describing the camera's location with respect to the robot's base $\left ( T_C^B \right )$. To estimate this transformation, an automated hand-eye calibration using an ArUco marker attached to the shaft of the PSM tool was used \cite{dvrk_camera_registration}. This calibration enabled us to determine the relative transformation between the camera and the robot base, facilitating an accurate alignment between the physical and virtual systems.

Lastly, it was required to identify the robot's location $\left ( T_E^B \right )$ with respect to its working environment. To calculate this transformation, we developed a module to register external CAD models or volumetric data from CT scans within the digital twin using the PSM tooltip as a digitizing device.
This module enabled us to register the environment model to the robotic instruments and endoscope.
In future work, the stereo endoscope could be used for 3D reconstruction of the environment, rather than relying on an \'{a} priori model, assuming the environment will not deform.

\subsection{Control strategies}
\label{sec:control}
When the communication state is normal, the input of the MTM is processed into a Cartesian goal, which is sent to both the remote robot and the digital twin. This was easily achieved by ensuring that the physical and digital robots followed the Collaborative Robotics Toolkit (CRTK) convention \cite{su2020collaborative}, a standardized set of commands and measurements for robotic systems, such as the {\tt servo\_cp} command observed in panel~(a) of Fig.~\ref{fig:main_figure}. Using CRTK as a common control interface simplifies the simultaneous control of the real and virtual robots and allows using any CRTK-compatible teleoperation device, such as a 6 DoF joystick or a small tabletop robot arm, to control the robots. Code for setting up the digital twin in AMBF and performing simultaneous control can be found in \cite{dvrk_digital_twin_teleoperation}.

When communication is interrupted, we implement two control strategies to illustrate how our digital twin framework can be used to evaluate the effectiveness of different strategies. From this point on, we will refer to these two strategies as the \textit{baseline} and the \textit{replay} strategies. For the \textit{baseline} strategy, the MTM is locked in place when communication is lost, and consequently no movement from either PSM or MTM is allowed. Once communication is restored, the MTM resumes normal function, granting the user teleoperation control again. This control scheme would be experienced in a system where there are no strategies to mitigate communication losses. 


For the \textit{replay} strategy, we aim to provide a smoother experience to the user by switching teleoperation to the digital twin robot during communication outages and then having the remote robot catch up with the digital twin upon restoring communication. In this case, when the communication is interrupted, the remote robot still remains stationary, but the digital twin is displayed as an AR overlay to enable the user to continue the task. During the outage, the user's commands are only sent to the digital twin and to a buffer as seen in panel~(b) of Fig.~\ref{fig:main_figure}. When communication is restored, the buffer is played back by skipping every other entry, making the remote robot follow the buffered user input at twice the speed (see panel~(c) in Fig.~\ref{fig:main_figure}). Note that the buffer continues to be appended with new commands during the recovery phase. For example, if the communication outage lasted for 1\,s, the system would need 0.5\,s to finish replaying the existing buffer, but during this time, new user input also enters the buffer. As a result, a total of 1\,s is required after the communication is restored to fully deplete the buffer and catch up to the user's input, before returning to normal teleoperation. Before the buffer is depleted, the AR overlay of the digital twin is shown, so that the operator would be able to continue their actions while observing the remote robot catch up. We design this recovery strategy specifically targeting relatively static scenes and short communication outages, 
and a different approach may be necessary in a highly dynamic environment and/or with long communication outages.

Some key benefits of the \textit{replay} strategy include eliminating disruptions in the user's workflow as the user is performing the task---the user can seamlessly transition to a virtual environment with the confidence that the system will catch up to them after the communication is restored. Moreover, in the safety-critical application of surgeries, this method provides the guarantee that the instrument still follows all the intended motions of the user and never performs an autonomous selection of actions.

\subsection{Digital twin implementation on the AMBF simulator}
\label{sec:ar}

\begin{figure}[!b]
    \centering
    \includegraphics[width=\linewidth]{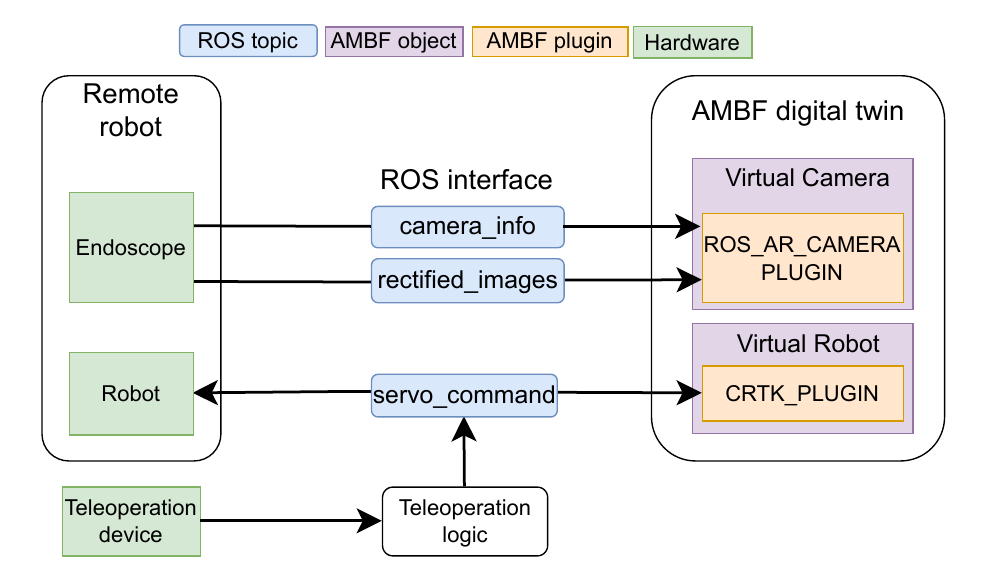}
    \caption{Software components of the digital twin system using AMBF simulator and ROS.}
    \label{fig:software_components}
\end{figure}

After performing the camera and environment calibration, the digital twin framework was implemented using AMBF as a simulation environment and ROS as a communication channel (see Fig.~\ref{fig:software_components}). To implement custom functionality in AMBF, we leveraged its plugin system which enables compiled libraries to be attached to any AMBF object for additional runtime functionalities. Specifically, two plugins were developed for the digital twin, one to overlay the virtual robot on the endoscopic images, and the other to create the CRTK communication interface.

The first plugin, \texttt{ROS\_AR\_CAMERA\_PLUGIN}, was attached to the simulator cameras to render virtual objects on top of endoscopic images. At initialization, this plugin subscribes to a ROS topic containing the calibration parameters and then modifies the projection matrix of the OpenGL camera inside AMBF based on these parameters. At runtime, the plugin receives images from the camera and uses them as a texture that is applied to the virtual world's background. This enables overlaying the digital twin on the endoscopic view that is presented to the user. This plugin assumes that the endoscope has been previously calibrated and that parameters and images are streamed using ROS as described in Sec.~\ref{sec:dt}. The second plugin, \texttt{CRTK\_PLUGIN}, was attached to the simulated robot to create the CRTK interface that enables the virtual robot to be commanded with the same commands as the real robot. 

\section{Experiment}
\label{sec:experiment}
\subsection{Experiment setup}
\label{sec:exp_setup}

To evaluate the efficacy of our proposed system, we conducted a user study involving eight engineering students, with institutional review board approval (HIRB \#00000701). We selected a peg transfer task, a widely recognized surgical training exercise in laparoscopic surgery, where participants must move objects between pegs on a pegboard without using the board or pegs as support. The task requires participants to transfer all three pegs from left to right and back again using only the left MTM.

\begin{figure}[!b]
    \centering
    \includegraphics[width=0.8\linewidth]{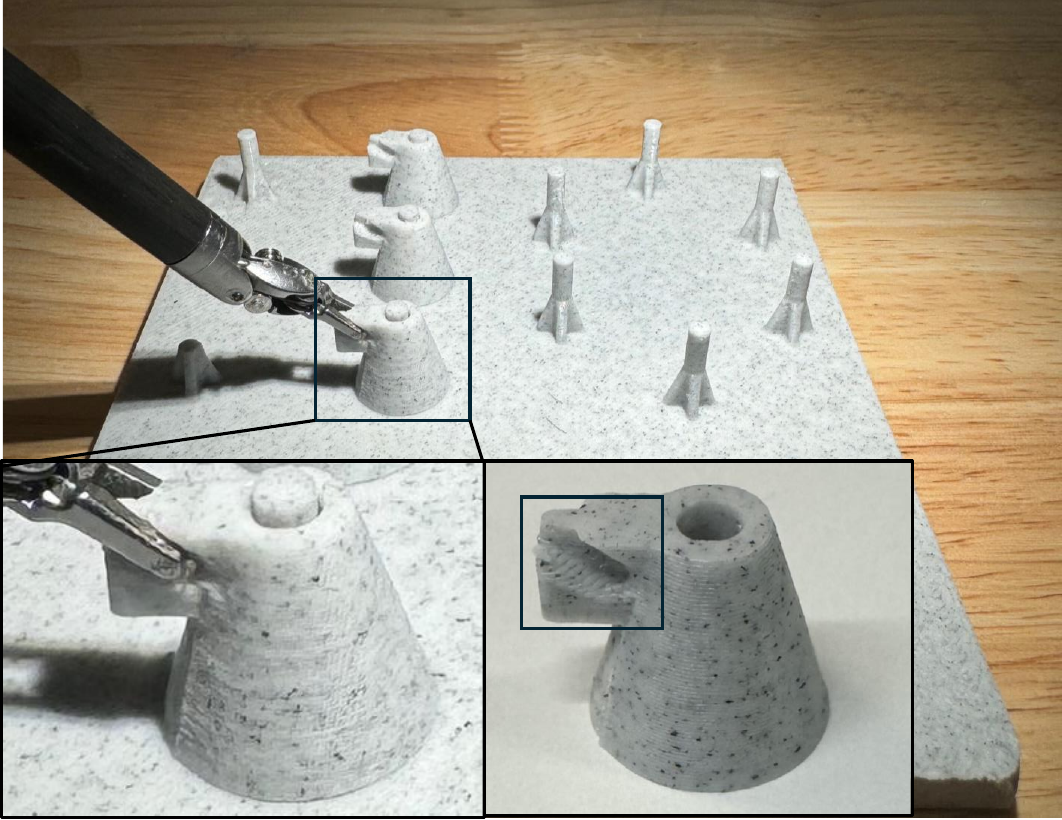}
    \caption{Peg and board design. Pegs were designed with a countersink to encourage the user to grasp the peg from the same position every time. The close-up picture shows the tool holding the peg from the intended grasping position.}
    \label{fig:peg}
\end{figure}

In these experiments, the environment consisted of a peg board and pegs. The peg board was assumed to be static; thus, rendering it as an AR overlay was unnecessary. On the other hand, the pegs changed position as they were grasped and released by the robotic gripper, and therefore they were included in the AR overlay. To obtain an accurate yet simple AR overlay of the virtual pegs, custom pegs were created with a countersink to encourage the users to grasp the peg from the same location each time (see Fig. \ref{fig:peg}). Additionally, grasping was determined through heuristics (e.g., when the gripper closes near a post). This allowed us to render the peg at a fixed location with respect to the gripper, producing realistic AR overlays that include the peg when appropriate during communication outage. In future work, this limitation will be avoided by using computer vision algorithms to estimate the pose of the pegs. 

Participants performed the task under the two developed control strategies: \textit{baseline} and \textit{replay}. In both conditions, participants experienced simulated, random periods of communication outage. Periods of normal communication have a mean of 3.2 seconds with a standard deviation of 0.15 seconds, while periods of communication outage have a mean of 0.8 seconds with a standard deviation of 0.1 seconds. These values were selected based on previous studies \cite{ishida2023semi, Nguan2008, Mersha2013}. Due to a system issue, we were unable to append commands to the buffer or display the AR overlay during the recovery phase, so only the buffered inputs from the communication outage period were replayed, and without an overlay.
These changes should only have a small negative impact on the users' experience, and thus we expect the results for the complete system to be better than the ones presented in \ref{sec:results}. Each user's time to fully complete the peg transfer task from the left to the right and then back is recorded, which includes both the duration of normal communication and the duration of communication outage. This task completion time serves as the first metric to compare between the two different conditions.

After completing the task, participants completed a NASA Task Load Index (TLX) survey and provided feedback on the control strategies used during the experiment. This feedback offers insight into the perceived usability and performance of each control condition and serves as our second set of metrics for comparing the conditions.

\subsection{Experimental results}
\label{sec:results}
The task completion time across all participants under the two different conditions is shown in Table~\ref{tab:time}. The mean completion time for the replay condition is $23.6\%$ lower than the mean completion time for the baseline condition, and a $t$-test shows statistical significance between the two experimental conditions, with $p < 0.005$.

\begin{figure}[ht]
    \centering
    \includegraphics[width=\linewidth]{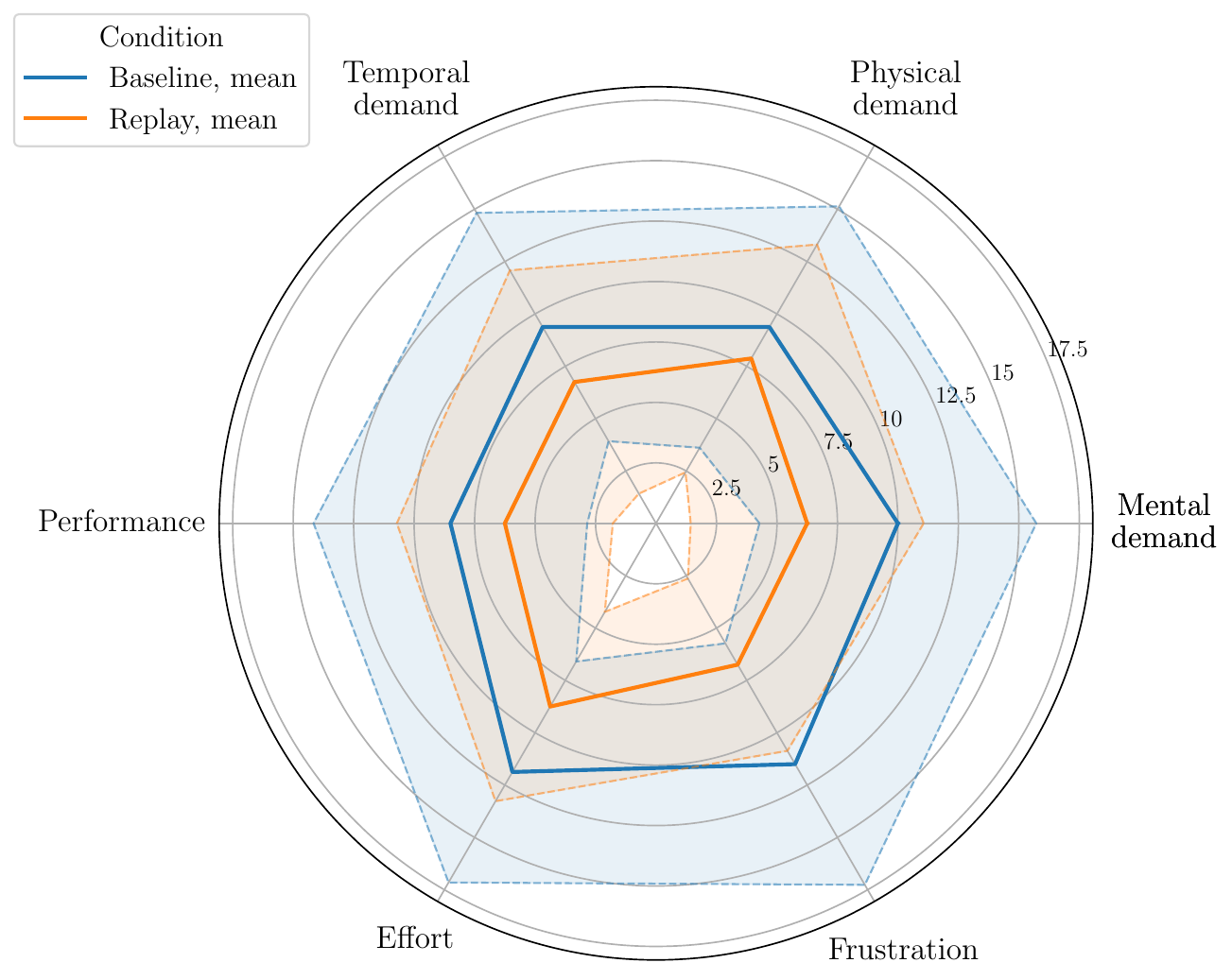}
    \caption{Radar plot with the mean and standard deviation on each dimension of the NASA Task Load Index (TLX), across all participants, for the two different conditions. The dotted lines with shaded area indicate the standard deviation for each condition and dimension. The values closer to the center indicate lower task load.}
    \label{fig:tlx}
\end{figure}

The mean and standard deviation of the NASA TLX survey result across all participants is shown in Fig.~\ref{fig:tlx}, separated by condition. The solid lines represent the means, and the dashed lines with the shaded area represent the regions within one standard deviation from the mean. Although the $t$-test does not show any statistical significance, the mean task load across all dimensions is still lower for the replay condition compared to the baseline condition.



\begin{table}[!b]
  \centering
  \caption{Task completion time (s) for all users, compared between baseline and replay strategies ($p<0.005$), as well as percent reduction comparing replay against baseline.}
  \begin{tabular}{c|cc|c}
    \specialrule{.15em}{0em}{.2em}
    \textbf{User} & \textbf{Baseline} & \textbf{Replay} & \textbf{Reduction (\%)} \\
    \specialrule{.05em}{.15em}{.15em} 
    1 & 140 & 75 & 46.5 \\
    2 & 212 & 157 & 25.9 \\
    3 & 274 & 217 & 20.9 \\
    4 & 198 & 191 & 3.4 \\
    5 & 139 & 136 & 2.0 \\
    6 & 182 & 101 & 44.4 \\
    7 & 147 & 103 & 29.9 \\
    8 & 137 & 116 & 15.4 \\
    \specialrule{.05em}{.15em}{.15em} 
    \textbf{Mean} & 178.6 & 137.0 & 23.6 \\
    \specialrule{.15em}{.1em}{0em}
  \end{tabular}
  \label{tab:time}
\end{table}


\section{Discussion}
\label{sec:discussion}
As shown in Sec.~\ref{sec:results}, the replay strategy yields consistent improvement in task performance when compared to the baseline against all metrics considered in this study. In particular, there is strong statistical significance demonstrating the reduction in task completion time. 

Interestingly, the mean reduction of 23\% in task completion time with replay is greater than the percentage of time under communication outage, which is 20\% (mean time under outage of 0.8\,s over mean total cycle length of 4\,s). 
This was observed for 5 of the 8 subjects, who had mean reductions greater than 20\%.
One potential explanation for this effect is that preventing MTM motion during communication outage in the baseline condition disrupted the user's workflow, which required them to use a small amount of time to recover to their original course of action after communication was restored. In contrast, the use of the digital twin and the replay strategy could compensate for some of these task inefficiencies by promoting a user workflow with fewer disruptions. However, we also note that the replay strategy did not lead to a noticeable time reduction for 2 of the 8 subjects, indicating that they may not have obtained a benefit from that strategy.

Similar benefits for using the digital twin can also be observed from the NASA TLX results in Fig.~\ref{fig:tlx}. Although the $t$-test does not report significance, we do see a uniform decrease in the survey response numbers with the replay condition compared to the baseline, which indicates a lower task load in general. One of the largest drops is in the ``frustration'' dimension, which can be attributed to the fact that the user's motion with the MTM is often disrupted during the \textit{baseline} condition. On the contrary, the \textit{replay} condition eliminates this effect and replaces it with a smooth user experience, hence decreasing user frustration while operating the system.

\section{Conclusions and Future Work}
\label{sec:summary}
This paper introduces a digital twin system for the dVRK robot, and demonstrates its utility to provide assistance during communication outages in the context of telerobotic surgery.
One natural extension of our work is to explore different communication imperfection patterns, including time delays and communication failures, and their mitigation strategies. The proposed buffer/replay strategy, where the buffer is replayed at twice the original speed, represents a first step toward mitigation of communication outage. However, mitigating more complex communication failures could require a more sophisticated replay mechanism, such as varying the replay speed so that slow motions are increased more than fast motions.

Alternatively, one could consider mitigation strategies that involve autonomous PSM motion during communication outage, and upon reestablishing communication, remedy the discrepancy between the autonomous motion and the user input. For instance, privileged information from the digital twin could benefit the distillation of some teacher-student reinforcement learning policy that can predict a short open-loop action sequence given the history of motion up until communication loss, similar to the approach in \cite{wang2024lessons}. For the lengths of communication outage considered in this paper, rolling out these trajectories open-loop when there is no communication could be a reliable prediction of user input.

Moreover, our approach was only tested in a relatively static and known environment. To tackle more realistic surgical tasks, it is critically important to model the environment using available sensors such as RGBD cameras. This would enable the system to modify the buffered motions based on physical abnormalities not accurately simulated in the digital twin or adjustments due to motions near high-risk areas of operation. Overall, rather than replaying the raw motions, the system should identify the user's intended actions during periods of communication outage and adapt buffered motions based on the environment model to ensure safe execution.



\section*{Acknowledgments}
The authors would like to express their appreciation for Haoying Zhou and Adnan Munawar for their support and feedback. H. Ishida was supported by NIDCD K08 Grant DC019708, by a research agreement with the Hong Kong Multi-Scale Medical Robotics Centre and by Johns Hopkins University internal funds.

\bibliography{teleop_under_comm_loss}
\bibliographystyle{IEEEtran}

\end{document}